\documentclass[a4paper]{article}

\usepackage[preprint]{INTERSPEECH_v2}
\usepackage[utf8]{inputenc}
\usepackage{cite}
\usepackage{multirow}

\DeclareFixedFont{\ttb}{T1}{txtt}{bx}{n}{8} 
\DeclareFixedFont{\ttm}{T1}{txtt}{m}{n}{8}  

\usepackage{color}
\definecolor{deepblue}{rgb}{0,0,0.5}
\definecolor{deepred}{rgb}{0.6,0,0}
\definecolor{deepgreen}{rgb}{0,0.5,0}

\usepackage{listings}

\newcommand\pythonstyle{\lstset{
    language=Python,
    basicstyle=\ttm,
    keywordstyle=\ttb\color{deepblue},
    emph={mbr,range,sum,zip},
    emphstyle=\ttb\color{deepred},
    stringstyle=\color{deepgreen},
    frame=tb,
    showstringspaces=false,
}}

\newcommand{\prob}{\mathbb{P}}
\newcommand{\expect}{\mathbb{E}}
\newcommand{\grad}{\frac{\partial}{\partial z}}

\newcommand{\sref}[1]{\S\ref{#1}}
\newcommand{\tabref}[1]{Table~\ref{#1}}
\newcommand{\figref}[1]{Figure~\ref{#1}}
\newcommand{\Eqref}[1]{(\ref{#1})}

\title{Optimizing expected word error rate via sampling for speech recognition}
\name{Matt Shannon}
\address{Google Inc., USA}
\email{mattshannon@google.com}

\toappear{IN PROCEEDINGS OF INTERSPEECH 2017}

\begin{document}

\maketitle
\begin{abstract}
State-level minimum Bayes risk (sMBR) training has become the de facto standard
for sequence-level training of speech recognition acoustic models.
It has an elegant formulation using the expectation semiring, and gives large
improvements in word error rate (WER) over models trained solely using
cross-entropy (CE) or connectionist temporal classification (CTC).
sMBR training optimizes the expected number of frames at which the reference and
hypothesized acoustic states differ.
It may be preferable to optimize the expected WER, but WER does not interact well
with the expectation semiring, and previous approaches based on computing expected
WER exactly involve expanding the lattices used during training.
In this paper we show how to perform optimization of the expected WER by sampling
paths from the lattices used during conventional sMBR training.
The gradient of the expected WER is itself an expectation, and so may be
approximated using Monte Carlo sampling.
We show experimentally that optimizing WER during acoustic model training gives
$5\%$ relative improvement in WER over a well-tuned sMBR baseline on a $2$-channel
query recognition task (Google Home).
\end{abstract}

\section{Introduction}
\label{sec:intro}
\emph{Minimum Bayes risk (MBR)} training
\cite{kaiser2000novel,povey2002minimum,doumpiotis2004pinched}
has been shown to be an effective way to train neural net--based acoustic models
and is widely used for state-of-the-art speech recognition systems
\cite{kingsbury2009lattice,sak2015fast,bellegarda2016state}.
MBR training minimizes an expected loss, where the loss measures the distance
between a reference and a hypothesis.
\emph{State-level MBR (sMBR)} training
\cite{gibson2006hypothesis,povey2007evaluation}
is a popular form of MBR training which uses a loss defined at the frame level
based on clustered context-dependent phonemes or subphonemes.
It is computationally tractable, has an elegant lattice-based formulation using
the \emph{expectation semiring} \cite{eisner2001expectation,heigold2008modified},
and has been shown to perform favorably in terms of \emph{word error rate (WER)}
compared to alternative losses that have been proposed
\cite{povey2007evaluation,vesely2013sequence}.

Given the prevalence of word error rate as an evaluation metric, it is natural to
consider using it during MBR training.
We refer to this as \emph{word-level edit-based MBR (word-level EMBR)} training.
However expected word edit distance over a lattice is harder to compute than the
expected frame-level loss used by sMBR training.
As a result, many approximations to the true edit distance have been proposed for
lattice-based MBR training
\cite{povey2002minimum,zheng2005improved,macherey2005investigations,
    gibson2006hypothesis,povey2007evaluation,yan2008minimum,gibson2008minimum-thesis}.
Gibson provides a systematic review of the strengths and weaknesses of various
approximations \cite[chapter 6]{gibson2008minimum-thesis}.
Heigold et al.\ show that it is possible to exactly compute expected word edit
distance over a lattice by first expanding the lattice using an approach sometimes
termed \emph{error marking} \cite{heigold2005minimum,van2015annotating}.
While impressive, error marking is computationally and implementionally non-trivial,
and increases the size of the lattices.
The memory required for FST determinization during error marking increases rapidly
with utterance length, and van Dalen and Gales found a practical limit of around
10 words with the conventional determinization algorithm or around 20 words with
an improved, memory-efficient algorithm \cite{van2015annotating}.
The time complexity also means error marking is unlikely to be compatible with
on-the-fly lattice generation.

In this paper we propose an alternative approach to lattice-based word-level EMBR
training.
The gradient of the expected loss optimized by MBR training may itself be written
as an expectation, allowing the gradient to be approximated by sampling.
The samples are paths through the lattice used during conventional sMBR training.
This approach is extremely flexible as it places almost no restriction on the form
of loss that may be used.
We use this approach to perform word-level EMBR training for two speech recognition
tasks and show that this improves WER compared to sMBR training.
To our knowledge this is also the first work to investigate word-level EMBR training
for state-of-the-art acoustic models based on neural nets.

Similar forms of sampled MBR training have been proposed previously.
Graves performed sampled EMBR training for the special case of a CTC model, noting
that the special structure of CTC allows trivial sampling and specialized approaches
to reduce the variance of the samples \cite{graves2014towards}.
Speech recognition may be viewed as a simple reinforcement learning problem where
an action consists of outputting a given word sequence.
In this view MBR training is learning a stochastic policy, and the sampling-based
approach described here is very similar to the \emph{REINFORCE} algorithm
\cite{williams1992simple,ranzato2016sequence}, though here we sample from a probability
distribution which is globally normalized instead of locally normalized, and there
are differences in the form of variance reduction used.

It should be noted that some previous investigations have reported better WER from
optimizing a state-level or phoneme-level criterion than a word-level criterion
\cite{povey2002minimum} \cite[chapter 7]{gibson2008minimum-thesis}.
However some have reported the opposite \cite{yan2008minimum}.
We find the word-level criterion more effective in our experiments but do not
investigate this question systematically.

In the remainder of this paper we review the sequence-level probabilistic model
used by MBR training (\sref{sec:fst-prob} and \sref{sec:seq-model}), describe
MBR training (\sref{sec:mbr-training}), discuss EMBR training
(\sref{sec:embr-training}), describe our sampling-based approach to MBR training
(\sref{sec:sampled-mbr}), and describe our experimental results (\sref{sec:expt}).


\section{FST-based probabilistic models}
\label{sec:fst-prob}
In this section we review how a weighted \emph{finite state transducer (FST)}
\cite{mohri2002weighted} may be globally normalized to obtain a probabilistic
model.
The model used for MBR training is defined in this way.

A weighted FST defined over the real semiring (also called the probability
semiring) may naturally be turned into a probabilistic model over a sequence of
its output labels \cite{eisner2002parameter}.
Each edge $e$ in the weighted FST has a real-valued weight $w_e \geq 0$ and an
optional output label taken from a specified output alphabet, and an optional
input label taken from a specified input alphabet.
The absence of an input or output label is usually denoted with a special epsilon
label.
A path $\pi$ through the FST from the initial state to the final state%
\footnote{%
    For simplicity we assume throughout that there is a single final state with
    trivial final weight and no edges leaving it.
}
consists of a sequence of edges, and the weight of the path $w(\pi)$ is defined
as the product $\prod_{e \in \pi} w_e$ of the weights of its edges.
The probability of a path is defined by globally normalizing this weight function:
\begin{equation}
    \label{eq:path-dist-general}
    \prob(\pi) = \frac{w(\pi)}{\sum_\pi w(\pi)}
\end{equation}
This is a Markovian distribution.
The output label sequence $\overline{y}(\pi)$ associated with a path $\pi$ is the
sequence of non-epsilon output labels along the path.
The probability of a sequence $y$ of output labels is defined as the sum of the
probabilities of all paths consistent with $y$:
\begin{equation}
    \label{eq:label-dist-general}
    \prob(y) = \sum_{\pi : \overline{y}(\pi) = y} \prob(\pi)
\end{equation}

\section{Sequence-level model for recognition}
\label{sec:seq-model}
In this section we briefly recap a conditional probabilistic model often used for
speech recognition.
This probabilistic model is an important component in sMBR training.

An \emph{acoustic model} specifies a mapping $z = \overline{z}(x, \lambda)$
from an \emph{acoustic feature vector sequence} $x = [x_t]_{t=1}^T$ to an
\emph{acoustic logit sequence} $z = [z_t]_{t=1}^T$ given model parameters $\lambda$.
Each dimension $q$ of the logit vector $z_t = [z_{t q}]_{q=1}^Q$ is typically
associated with a cluster of context-dependent phonemes or sub-phonemes
\cite{dahl2012context}.
In this paper the acoustic model is a stacked \emph{long short-term memory (LSTM)}
network \cite{hochreiter1997long}.
Given a logit sequence $z$, we define a weighted FST $U(z)$ as the composition
\begin{equation}
    U(z) = S(z) \circ (C \circ L \circ G)
\end{equation}
of a \emph{score FST} $S(z)$ and a \emph{decoder graph FST} $C \circ L \circ G$.
The decoder graph is itself a composition and incorporates weighted information
about context-dependence (C), pronunciation (L) and word sequence plausibility (G)
\cite{mohri2008speech}.
The score FST input and output alphabets and the decoder graph input alphabet are
$\{1, \ldots, Q\}$.
The decoder graph output alphabet is the vocabulary.
We refer to $U(z)$ as the \emph{unrolled decoder graph} since composition with
$S(z)$ effectively unrolls the decoder graph over time.
The score FST $S(z)$ has a simple ``sausage'' structure with $T + 1$
states and an edge from state $t$ to state $t + 1$ with input and output label $q$
and log weight $z_{t q}$ for each frame $t$ and cluster index $q$.

The conditional probabilistic model $\prob(y | x, \lambda)$ over a word sequence
$y = [y_j]_{j=1}^J$ given an acoustic feature vector sequence $x = [x_t]_{t=1}^T$
is obtained by applying the procedure described in \sref{sec:fst-prob} to the
unrolled decoder graph $U(\overline{z}(x, \lambda))$.
This gives
\begin{equation}
    \label{eq:path-dist}
    \prob(\pi | x, \lambda) = \frac{w(\pi, z)}{\sum_\pi w(\pi, z)}
\end{equation}
where $z = \overline{z}(x, \lambda)$ and $w(\pi, z)$ is the weight of a path.

It will be helpful to know the gradient of the log weight and log probability with
respect to the acoustic logits.
Due to the way the score FST is constructed, the sequence of non-epsilon input
labels encountered along a path through $U(z)$ consists of a cluster index $q_t$
for each frame $t \in \{1, \ldots, T\}$.
The overall log weight $\log w(\pi, z)$ has an additive contribution
$\sum_t z_{t q_t}$ from the acoustic model, and the gradient $\grad\log w(\pi, z)$
consists of a $T \times Q$ matrix which has a one at $(t, q_t)$ for each frame $t$
and zeros elsewhere.
The gradient of the log probability is given by
\begin{equation}
    \label{eq:log-prob-grad}
    \grad \log \prob(\pi | z) = \grad \log w(\pi, z) - \expect \grad \log w(\pi, z)
\end{equation}

\section{Minimum Bayes risk (MBR) training}
\label{sec:mbr-training}
\emph{Minimum Bayes risk (MBR)} training
\cite{kaiser2000novel,povey2002minimum,doumpiotis2004pinched}
minimizes the expected loss%
\footnote{%
    MBR training is named by analogy with MBR decoding \cite{goel2000minimum},
    but otherwise has little to do with Bayesian modeling or decision theory.
}
\begin{equation}
    \expect L(y, y_\text{ref})
        = \sum_\pi \prob(\pi | x, \lambda) L(\overline{y}(\pi), y_\text{ref})
\end{equation}
as a function of $\lambda$, where the loss $L(y, y_\text{ref})$ specifies how bad
it is to output $y$ when the reference word sequence is $y_\text{ref}$.
Broadly speaking MBR concentrates probability mass:
a sufficiently flexible model trained to convergence with MBR will assign a
probability of $1$ to the word sequence(s) with smallest loss.

\emph{State-level minimum Bayes risk (sMBR)} training
\cite{gibson2006hypothesis,povey2007evaluation}
defines the loss $L(\pi, \pi_\text{ref})$, now over $\pi$ instead of $y$, to be
the number of frames at which the current cluster index differs from the reference
cluster index.
The sequence of time-aligned reference cluster indices is typically obtained
using forced alignment.
This loss has a number of disadvantages compared to minimizing the number of
word errors \cite[chapter 6]{gibson2008minimum-thesis}.
However it has the advantage that the expected loss and its gradient may be
computed tractably using an elegant formulation based on the
\emph{expectation semiring} \cite{eisner2001expectation,heigold2008modified}.
The crucial property of the loss $L$ that makes it tractable is that the loss of
a path can be decomposed additively over the edges in the path.

Typically it is not feasible to perform sMBR training over the full unrolled
decoder graph $U(z)$, and a lattice containing a subset of paths is used
instead
\cite{valtchev1996lattice,kingsbury2009lattice}.
There has been some interest recently in performing exact ``lattice-free''
sequence training using a simplified decoder graph \cite{povey2016purely}.

\section{Edit-based MBR (EMBR) training}
\label{sec:embr-training}
Given the prevalence of \emph{word error rate (WER)} as an evaluation metric,
a natural loss to use for MBR training is the edit distance between the
reference and hypothesized word sequences.
We refer to MBR training using a Levenshtein distance as the loss as
\emph{edit-based MBR (EMBR)} training, and to using the number of word errors as
\emph{word-level EMBR} training.
The number of word errors is the result of a dynamic programming computation and
does not decompose additively over the edges in a path, meaning the expectation
semiring approach cannot be applied without modification.
As discussed in \sref{sec:intro}, it is in fact possible to exactly compute the
expected number of word errors over a lattice by expanding the lattice using
error marking, but this approach produces larger lattices, in practice limits the
maximum utterance length, and is not compatible with on-the-fly lattice generation
\cite{heigold2005minimum,van2015annotating}.

\section{Sampled MBR training}
\label{sec:sampled-mbr}
In this section we look at a simple sampling-based approach to computing the MBR
loss value and its gradient.
This approach is essentially agnostic to the form of loss used, allowing us to
perform EMBR training simply and efficiently.
We consider the expected loss as a function of $z$ rather than $\lambda$ since
this provides the gradients required to implement sampled MBR in a modular
graph-of-operations framework such as TensorFlow \cite{abadi2016tensorflow}.

Samples from $\prob(\pi | z)$ can be drawn efficiently using
\emph{backward filtering--forward sampling} \cite{loeliger2009estimating}.
First the conventional backward algorithm is used to compute the sum $\beta_i$ of
the weights of all partial paths from each FST state $i$ to the final state.
The FST is then reweighted using $\beta$ as a \emph{potential function},
i.e.\ replacing the weight $w_e$ of each edge $e$, which goes from some state $i$
to some state $j$, with $\beta_i^{-1} w_e \beta_j$ \cite{mohri2001weight}.
This results in an equivalent FST which is locally normalized or \emph{stochastic},
i.e.\ the sum of edge weights leaving each state is one \cite{mohri2001weight}.
Samples from the reweighted FST can be drawn using simple ancestral sampling,
i.e.\ sample an edge leaving the initial state, then an edge leaving the end state
of the sampled edge, and repeat until we reach the final state.
In our implementation the $\beta$ values are computed once to draw multiple samples
and the reweighting is performed on-the-fly during sampling.

We can approximate the expected loss using a straightforward Monte Carlo
approximation:
\begin{align}
    \expect L(y, y_\text{ref})
        &= \sum_\pi \prob(\pi | z) L(\overline{y}(\pi), y_\text{ref})
    \\
    \label{eq:monte-carlo-value}
        &\approx \frac{1}{I} \sum_{i=1}^I L_i = \overline{L_i}
\end{align}
where each $\pi_i$ is an independent sample from $\prob(\pi | z)$ and
$L_i = L(\overline{y}(\pi_i), y_\text{ref})$.
We write $\overline{L_i}$ to denote averaging over samples.

The true gradient, writing $L(\pi)$ for $L(\overline{y}(\pi), y_\text{ref})$, is
\begin{align}
    \grad \expect L(\pi)
        &= \grad \sum_\pi \prob(\pi | z) L(\pi)
    \\
        &= \sum_\pi \prob(\pi | z) L(\pi) \grad\log\prob(\pi | z)
\end{align}
This can be re-expressed using \Eqref{eq:log-prob-grad} as
\begin{equation}
    \label{eq:sampled-wmbr}
    \expect \left[ L(\pi) \grad\log w(\pi, z) \right] -
        \expect L(\pi) \expect \left[ \grad\log w(\pi, z) \right]
\end{equation}
Thus the MBR gradient is a scalar-vector covariance \cite{heigold2008modified},
i.e.\ it is of the form $\expect [x y] - \expect x \expect y$, and an unbiased
estimate is
\begin{equation}
    \label{eq:monte-carlo-gradient}
    \grad \expect L(\pi)
        \approx \frac{I}{I - 1} \overline{(L_i - \overline{L_i}) \grad\log w(\pi_i, z)}
\end{equation}
We refer to using \Eqref{eq:monte-carlo-gradient} during gradient-based training as
\emph{sampled MBR training}.
It has the intuitive interpretation that samples with worse-than-average loss have
the log weight of the corresponding path reduced, and samples with better-than-average
loss have the log weight of the corresponding path increased.
The subtraction of $\overline{L_i}$ in \Eqref{eq:monte-carlo-gradient} makes the
estimated gradient invariant to an overall additive shift in loss, and may be seen
as performing variance reduction%
\footnote{%
    Sampled MBR without variance reduction may be defined by approximating
    $\expect L(\pi) \grad\log\prob(\pi | z)$ as
    $\overline{L_i \grad\log\prob(\pi_i | z)}$.
    The variance reduced version we have presented is equivalent to approximating
    $\expect L(\pi) \grad\log\prob(\pi | z) -
        \expect L(\pi) \expect \grad\log\prob(\pi | z)$
    as
    $\tfrac{I}{I - 1} \left( \overline{L_i \grad\log\prob(\pi_i | z)} -
        \overline{L_i} \overline{\grad\log\prob(\pi_i | z)} \right)$.
    Both are unbiased estimates of the desired gradient since
    $\expect \grad\log\prob(\pi | z) = 0$.
},
often regarded as extremely important in sampling-based policy optimization for
reinforcement learning
\cite{sutton1999policy,sehnke2010parameter,mnih2014recurrent}.

Example code for sampled MBR training is shown in \figref{fig:sampled-mbr-code}.
Our implementation consisted of a TensorFlow op with input acoustic logits and
output expected loss.
\begin{figure}
    \pythonstyle
\begin{lstlisting}[language=Python]
def mbr(sample_path, collapse_path, get_gammas,
        get_loss, num_samples, ref_labels):
  paths = [sample_path()
           for _ in range(num_samples)]
  losses = [
      get_loss(collapse_path(path), ref_labels)
      for path in paths
  ]
  mean_loss = np.mean(losses)
  mean_gammas = sum([
      get_gammas(path) * (loss - mean_loss)
      for path, loss in zip(paths, losses)
  ]) / (num_samples - 1)
  return mean_loss, mean_gammas
\end{lstlisting}
    \caption{%
        Example python implementation of sampled MBR training.
        Here \texttt{sample\_path} returns a sampled path from a lattice,
        \texttt{collapse\_path} ($\overline{y}$ in \Eqref{eq:label-dist-general})
        takes a path to a word sequence,
        \texttt{get\_gammas} takes a path to a $T \times Q$ matrix with a one
        for each (frame, cluster index) which occurs in the path
        (see \sref{sec:seq-model}), and
        \texttt{get\_loss} computes Levenshtein distance in the case of EMBR
        training.
    }
    \label{fig:sampled-mbr-code}
\end{figure}

\section{Experiments}
\label{sec:expt}
We compared sMBR and sampled word-level EMBR training for two model architectures
on two speech recognition tasks.

\subsection{CTC-style 2-channel Google Home model}
We compared the performance of sMBR and EMBR training for a 2-channel query
recognition task (Google Home) using a CTC-style model where the set of clustered
context-dependent phonemes includes a blank symbol.

Our acoustic feature vector sequence was $480$-dimensional with a $30\,\text{ms}$
frame step, and consisted of a stack of three $10\,\text{ms}$ frames of $80$ log
mel filterbanks for each channel.
A stacked unidirectional LSTM acoustic model with $5$ layers of $700$ cells each
was used, followed by a linear layer with output dimension $8192$ (the number of
clusters).
These ``raw'' logits were post-processed by applying log normalization with a
softmax-then-log then adding $-1.95$ to the logit for CTC blank and scaling the
logits for the remaining clusters by $0.5$ to mimic what is often done during
decoding for CTC models; this post-processing does not seem to be critical for
good performance.
The initial CTC system was trained from a random initialization using $739$ million
steps of asynchronous stochastic gradient descent, using context-dependent
phonemes as the reference label sequence.
The sMBR and EMBR systems were trained for a further $900$ million steps, with the
best systems selected by WER on a small held-out dev set being obtained after a
total of roughly $1035$ million and $1569$ million steps respectively.
For both CTC and MBR training, each step computed the gradient on one whole
utterance.
Lattices for MBR training were generated on-the-fly.
Alignments for sMBR training were computed on-the-fly.
The decoder graph used during MBR training was constructed from a weak bigram
language model estimated on the training data, and a CTC-style context-dependent
C transducer with blank symbol was used \cite{sak2015learning}.
EMBR training used $100$ samples per step.
The WER used during EMBR training was computed with language-specific capitalization
and punctuation normalization.
The learning rate for EMBR training was set $5$ times larger than for sMBR training
because the number of word errors is typically smaller than the number of frame
errors, and the typical gradient norm values during sMBR and EMBR training reflected
this.
We verified that using a $5$ times larger learning rate for sMBR was detrimental.
The training data was a voice search--specific subset of a corpus of around $20$
million anonymized utterances with artificial room reverbation and noise added.
WER was evaluated on three corpora of anonymized $2$-channel Google Home utterances.

\begin{figure}[t]
    \includegraphics[width=\linewidth]{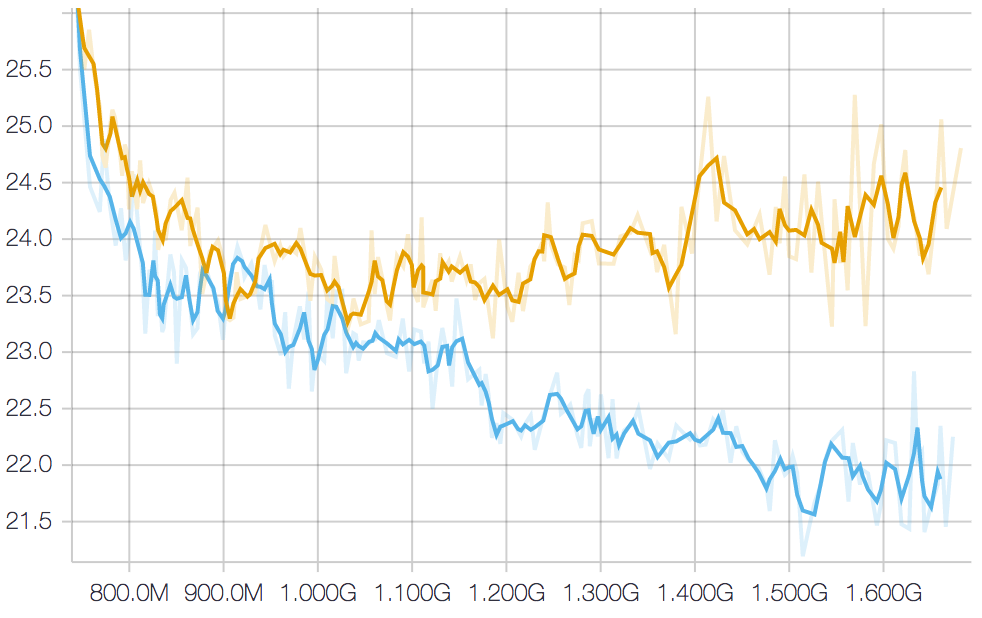}
    \caption{%
        WER computed during training on the dev set for Google Home for sMBR (top)
        and word-level EMBR (bottom).
    }
    \label{fig:home-trainer-wer}
\end{figure}

WER on the dev set during training is presented in \figref{fig:home-trainer-wer}.
Somewhat contrary to common wisdom that sMBR training converges quickly, we
observe WER continuing to improve for around $300$ million steps.
After $1$ billion total steps, sMBR training gets worse on the dev set and stays
roughly constant on the training set (not shown), meaning it is suffering from
overfitting.
EMBR training takes a long time to achieve its full gains; indeed it is not clear
it has converged even at $1.6$ billion total steps.
EMBR performance appears to be as good or better than sMBR after any number of
steps.
An EMBR step with $100$ samples in our set-up took around the same time as an
sMBR step.
EMBR has little overhead because sampling paths given the beta probabilities is
cheap and because overall computation time for both sMBR and EMBR is typically
dominated by the stacked LSTM computations and lattice generation.

The evaluation results are presented in \tabref{tab:home-results}.
We can see the sampled word-level EMBR training gave a 4--5 \% relative gain
over sMBR training.
\begin{table}[th]
    \centering
    \begin{tabular}{ l c c c c }
    \toprule
        \multirow{2}{*}{Test set} & \multirow{2}{*}{Num utts} &
            \multicolumn{2}{ c }{WER} & \multirow{2}{*}{Relative} \\
        \cline{3-4}
        & & sMBR & WEMBR & \\
    \midrule
        Home 1 & 21k & 6.5 & 6.2 & -5\% \\
        Home 2 & 22k & 6.3 & 5.9 & -6\% \\
        Home 3 & 22k & 6.9 & 6.6 & -4\% \\
    \midrule
        Overall & 65k & 6.6 & 6.2 & -5\% \\
    \bottomrule
    \end{tabular}
    \caption{%
        Word error rates for CTC-style 2-channel Google Home model trained with
        sMBR and word-level EMBR.
    }
    \label{tab:home-results}
\end{table}

\subsection{Non-CTC 1-channel voice search model}
We also compared the performance of sMBR and EMBR training for a 1-channel voice
search task.
Our acoustic feature vector sequence was $512$-dimensional with a $30\,\text{ms}$
frame step, and consisted of a stack of four $10\,\text{ms}$ frames of $128$ log
mel filterbanks.
The stacked LSTM had $5$ layers of $600$ cells each.
An initial cross-entropy system was trained for $305$ million steps.
The sMBR and EMBR systems were trained for a futher $420$ million steps, and the
models with best WER on a held-out dev set were at around $680$ million and $720$
million total steps respectively.
The systems were trained on a corpus of around $15$ million anonymized voice search
and dictation utterances with artificial reverb and noise added and evaluated on
four voice search test sets including one with noise added.

\begin{figure}[t]
    \includegraphics[width=\linewidth]{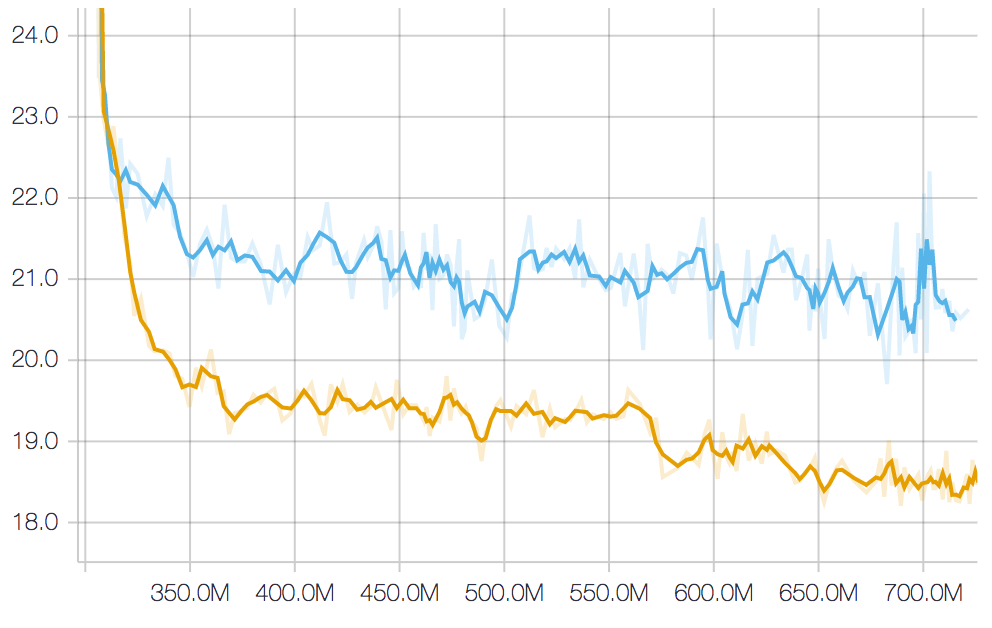}
    \caption{%
        WER computed during training on the dev set for voice search for sMBR
        (top) and word-level EMBR (bottom).
    }
    \label{fig:lfr-trainer-wer}
\end{figure}

WER on the dev set during training is presented in \figref{fig:lfr-trainer-wer},
and results are presented in \tabref{tab:lfr-results}.
In these experiments we used the same learning rate for sMBR and EMBR training
since we were not yet fully aware of the mismatch in dynamic range; using a smaller
learning rate may improve sMBR performance.
We observed consistent gains from word-level EMBR training on this task.
\begin{table}[th]
    \centering
    \begin{tabular}{ l c c c c }
    \toprule
        \multirow{2}{*}{Test set} & \multirow{2}{*}{Num utts} &
            \multicolumn{2}{ c }{WER} & \multirow{2}{*}{Relative} \\
        \cline{3-4}
        & & sMBR & WEMBR & \\
    \midrule
        VS 0 & 13k & 11.7 & 11.3 & -3\% \\
        VS 0 noisified & 13k & 14.0 & 13.2 & -6\% \\
        VS 1 & 15k & 9.4 & 9.0 & -4\% \\
        VS 2 & 22k & 10.7 & 10.2 & -5\% \\
    \bottomrule
    \end{tabular}
    \caption{%
        Word error rates for non-CTC 1-channel voice search model trained with
        sMBR and word-level EMBR.
    }
    \label{tab:lfr-results}
\end{table}

\section{Conclusion}
We have seen that sampled word-level EMBR training provides a simple and effective
way to optimize expected word error rate during training, and that this improves
empirical performance on two speech recognition tasks with disparate architectures.



\section{Acknowledgements}
Many thanks to Erik McDermott for his patient and helpful feedback on successive
drafts of this manuscript, and to Haşim Sak for many fruitful and enjoyable
discussions while implementing sMBR and EMBR training in TensorFlow.

\clearpage
\bibliographystyle{IEEEtran}
\bibliography{paper}

\end{document}